\documentclass{article}
\usepackage{colacl}
\usepackage{amsmath,epsfig,alltt,boxedminipage}

\title{\vspace*{-2ex}Annotation Graphs as a Framework for \\
  Multidimensional Linguistic Data Analysis}
\author{Steven Bird \and Mark Liberman\\
Linguistic Data Consortium, University of Pennsylvania\\
3615 Market St, Philadelphia, PA 19104-2608, USA\\
{\tt \{sb,myl\}@ldc.upenn.edu}}
\def\smtt#1{{\small\tt #1}}

\newenvironment{sv}{\scriptsize\begin{alltt}}{\end{alltt}\normalsize}
\newenvironment{bv}{\noindent\scriptsize\begin{boxedminipage}[t]{\linewidth}%
  \begin{alltt}}{\end{alltt}\end{boxedminipage}\normalsize\vspace*{1ex}}

\begin{document}

\maketitle
\vspace{-0.5in}
\begin{abstract}
In recent work we have presented a formal framework for
linguistic annotation based on labeled acyclic digraphs.
These `annotation graphs' offer a simple yet powerful
method for representing complex annotation structures
incorporating hierarchy and overlap.
Here, we motivate and illustrate our approach using
discourse-level annotations of text and speech data
drawn from the CALLHOME, COCONUT, MUC-7, DAMSL and TRAINS
annotation schemes.  With the help of domain specialists,
we have constructed a hybrid multi-level annotation for
a fragment of the Boston University Radio Speech Corpus
which includes the following levels:
segment, word, breath, ToBI, Tilt, Treebank, coreference
and named entity.
We show how annotation graphs can represent
hybrid multi-level structures which derive from a diverse
set of file formats.
We also show how the approach facilitates substantive
comparison of multiple annotations of a single signal
based on different theoretical models.
The discussion shows how annotation graphs
open the door to wide-ranging integration of tools, formats
and corpora.
\end{abstract}

\section{Annotation Graphs}
\label{sec:intro}

When we examine
the kinds of speech transcription and annotation found in many
existing `communities of practice', we see commonality of
abstract form along with diversity of concrete format.
Our survey of annotation practice
\cite{BirdLiberman99} attests to this commonality
amidst diversity.  (See [\smtt{www.ldc.upenn.edu/annotation}]
for pointers to online material.)
We observed that
all annotations of recorded linguistic signals require one unavoidable basic
action: to associate a label, or an ordered sequence of labels, with a
stretch of time in the recording(s). Such annotations also typically
distinguish labels of different types, such as spoken words vs.\ non-speech
noises. Different types of annotation often span different-sized
stretches of recorded time, without necessarily forming a strict
hierarchy: thus a conversation contains (perhaps overlapping)
conversational turns, turns contain (perhaps interrupted) words, and
words contain (perhaps shared) phonetic segments. Some types of annotation
are systematically incommensurable with others: thus disfluency structures
\cite{Taylor95}
and focus structures \cite{Jackendoff72} often
cut across conversational turns and syntactic constituents.

A minimal formalization of this basic set of practices is a directed
graph with fielded records on the arcs and optional time references on
the nodes.  We have argued that this minimal formalization in fact has
sufficient expressive capacity to encode, in a reasonably intuitive
way, all of the kinds of linguistic annotations in use today.  We
have also argued that this minimal formalization has good properties with
respect to creation, maintenance and searching of annotations. We
believe that these advantages are especially strong in the case of
discourse annotations, because of the prevalence of cross-cutting
structures and the need to compare multiple annotations representing
different purposes and perspectives.

Translation into annotation graphs does not magically create
compatibility among systems whose semantics are different.
For instance, there are many
different approaches to transcribing filled pauses in English -- each
will translate easily into an annotation graph framework, but their
semantic incompatibility is not thereby erased.
However, it does enable us to focus on the substantive differences
without having to be concerned with diverse formats, and without
being forced to recode annotations in an agreed, common format.
Therefore, we focus on the {\it structure} of annotations,
independently of domain-specific concerns about
permissible tags, attributes, and values.

As reference corpora are published for
a wider range of spoken language genres,
annotation work is increasingly reusing the same primary data.
For instance, the Switchboard corpus
[\smtt{www.ldc.upenn.edu/Catalog/LDC93S7.html}]
has been marked up for disfluency \cite{Taylor95}.
See [\smtt{www.cis.upenn.edu/\~{}treebank/switchboard-} \smtt{sample.html}]
for an example, which also includes a separate part-of-speech
annotation and a Treebank-style annotation.
\newcite{Hirschman97} give an example of MUC-7 coreference
annotation applied to an existing TRAINS dialog annotation
marking speaker turns and overlap.
We shall encounter a number of such cases here.

\subsection*{The Formalism}

\newtheorem{defn}{Definition}

As we said above, we take an annotation label to be a fielded
record.  A minimal but sufficient set of fields would be:

\begin{description}
\item[type] this represents a level of
  an annotation, such as the segment, word
  and discourse levels;
\item[label] this is a contentful property, such as
  a particular word, a speaker's name, or a discourse function;
\item[class] this is an optional field which permits
  the arcs of an annotation graph to be co-indexed as
  members of an equivalence class.\footnote{
  We have avoided using explicit pointers since we prefer not
  to associate formal identifiers to the arcs.  Equivalence classes
  will be exemplified later.}
\end{description} 

\noindent
One might add further fields for holding
comments, annotator id, update history, and so on.

Let $T$ be a set of types, $L$ be a set of labels, and $C$ be a set of classes.
Let $R$ = \(\{\left<t, l, c\right> \mid t\in T, l\in L, c\in C\}\), the set of
records over $T, L, C$.  Let $N$ be a set of nodes.
Annotation graphs (AGs) are now defined as follows:

\begin{figure*}
\centerline{\epsfig{figure=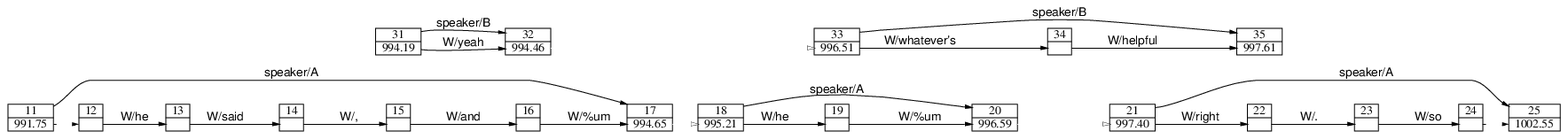,width=\linewidth}}
\caption{Graph Structure for LDC Telephone Speech Example}\label{fig:callhome}
\vspace*{2ex}\hrule\vspace*{2ex}
\centerline{\epsfig{figure=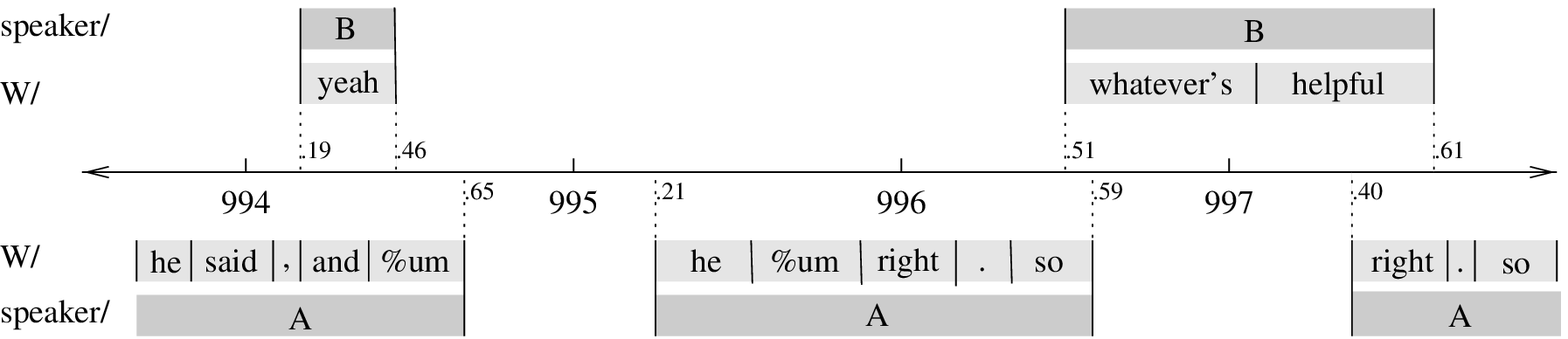,width=\linewidth}}
\caption{Visualization for LDC Telephone Speech Example}\label{fig:ldc2}
\vspace*{2ex}\hrule
\end{figure*}

\begin{defn}
An \textbf{annotation graph} $G$ over $R, N$
is a set of triples having the form
$\left<n_1, r, n_2\right>$, $r\in R$, $n_1, n_2\in N$, which
satisfies the following conditions:

\begin{enumerate}
\setlength{\itemsep}{0pt}
\item
\(
  \left<
    N,
    \left\{
      \left<n_1,n_2\right> \mid
      \left<n_1,r,n_2\right> \in A
    \right\}
  \right>
\)
is a labelled acyclic digraph.

\item $\tau: N \rightharpoonup \Re$
is an order-preserving map assigning times to (some of) the nodes.
\end{enumerate}
\end{defn}

For detailed discussion of these structures, see \cite{BirdLiberman99}.
Here we present a fragment (taken from Figure~\ref{fig:damsl} below)
to illustrate the definition.
For convenience the components of the fielded records which decorate
the arcs are separated using the slash symbol.  The example
contains two word arcs, and a discourse tag encoding `influence on speaker'.
No class fields are used.  Not all nodes have a time reference.
\vspace{2ex}

\centerline{\epsfig{figure=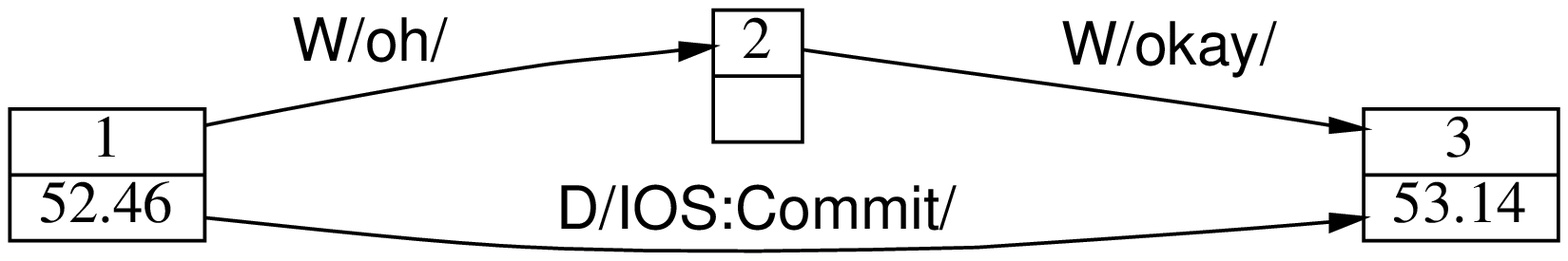,width=.9\linewidth}}

\noindent
The minimal annotation graph for this structure is as follows:
\vspace*{-3ex}

\def\z#1{\mbox{\smtt{#1}}}
\begin{eqnarray*}
T & = & \left\{\z{W}, \z{D}\right\} \\
L & = & \left\{\z{oh}, \z{okay}, \z{IOS:Commit}\right\} \\
C & = & \emptyset \\
N & = & \left\{1, 2, 3\right\} \\
\tau
  & = & \left\{\left<1, 52.46\right>, \left<3, 53.14\right> \right\} \\
A & = & \left\{
    \begin{array}{l}
      \left<1, \z{W/oh/}, 2\right>,\\
      \left<2, \z{W/okay/}, 3\right>,\\
      \left<1, \z{D/IOS:Commit/}, 3\right>
    \end{array}
  \right\}
\end{eqnarray*}

XML is a natural `surface representation' for annotation graphs and
could provide the primary exchange format.  A particularly simple
XML encoding of the above structure is shown below;
one might choose to use a richer XML encoding in practice.

\begin{bv}
<annotation>
  <arc>
    <begin id=1 time=52.46>
    <label type="W" name="oh">
    <end id=2>
  </arc>
  <arc>
    <begin id=2>
    <label type="W" name="okay">
    <end id=3 time=53.14>
  </arc>
  <arc>
    <begin id=1 time=52.46>
    <label type="D" name="IOS:Commit">
    <end id=3 time=53.14>
  </arc>
</annotation>
\end{bv}

\section{AGs and Discourse Markup}

\subsection{LDC Telephone Speech Transcripts}
\label{sec:callhome}

The LDC-published CALLHOME corpora include digital audio, 
transcripts and lexicons for telephone conversations in several
languages,
and are designed to support research on speech recognition
[\smtt{www.ldc.upenn.edu/Catalog/LDC96S46.html}].
The transcripts exhibit abundant overlap between speaker turns.
What follows is a typical fragment of an annotation.  Each stretch of
speech consists of a begin time, an end time, a speaker designation,
and the transcription for the cited stretch of time.
We have augmented the annotation with \smtt{+}
and \smtt{*} to indicate partial and total overlap (respectively) with
the previous speaker turn.

\begin{bv}
  962.68 970.21 A: He was changing projects every couple
    of weeks and he said he couldn't keep on top of it.
    He couldn't learn the whole new area  
* 968.71 969.00 B: %mm.  
  970.35 971.94 A: that fast each time.  
* 971.23 971.42 B: %mm.    
  972.46 979.47 A: %um, and he says he went in and had some
    tests, and he was diagnosed as having attention deficit
    disorder. Which
  980.18 989.56 A: you know, given how he's how far he's
    gotten, you know, he got his degree at &Tufts and all,
    I found that surprising that for the first time as an
    adult they're diagnosing this. %um  
+ 989.42 991.86 B: %mm. I wonder about it. But anyway.  
+ 991.75 994.65 A: yeah, but that's what he said. And %um  
* 994.19 994.46 B: yeah.  
  995.21 996.59 A: He %um  
+ 996.51 997.61 B: Whatever's helpful.  
+ 997.40 1002.55 A: Right. So he found this new job as a
    financial consultant and seems to be happy with that.  
  1003.14 1003.45 B: Good.  
\end{bv}

Long turns (e.g.\ the period from 972.46 to 989.56 seconds) were
broken up into shorter stretches for the convenience of the
annotators and to provide additional time references.
A section of this annotation which includes an example of total
overlap is represented in annotation graph form in
Figure~\ref{fig:callhome}, with the accompanying visualization
shown in Figure~\ref{fig:ldc2}.  (We have no commitment to this
particular visualization; the graph structures can be visualized in
many ways and the perspicuity of a visualization format will be
somewhat domain-specific.)

The turns are attributed to speakers using the \smtt{speaker/} type.
All of the words, punctuation and disfluencies are given the \smtt{W/}
type, though we could easily opt for a more refined version in which
these are assigned different types.  The class field is not used here.
Observe that each speaker turn is a disjoint piece of graph
structure, and that hierarchical organisation uses the
`chart construction' \cite[179ff]{GazdarMellish89}.
Thus, we make a logical distinction between the situation where
the endpoints of two pieces of annotation necessarily coincide
(by sharing the same node) from the situation where endpoints happen
to coincide (by having distinct nodes which contain the same time
reference).  The former possibility is required for hierarchical
structure, and the latter possibility is required for overlapping
speaker turns where words spoken by different speakers may happen
to sharing the same boundary.

\subsection{Dialogue Annotation in COCONUT}

\begin{figure*}
\begin{verbatim}
Accept, Commit              S1:   (a)  Let's take the blue rug for 250,
                                  (b)  my rug wouldn't match
Open-Option                       (c)  which is yellow for 150.
Action-Directive            S2:   (d)  we don't have to match...
Accept(d), Offer, Commit    S1:   (e)  well then let's use mine for 150
\end{verbatim}
\caption{Dialogue with COCONUT Coding Scheme}\label{fig:coconut-text}
\vspace*{2ex}\hrule\vspace*{2ex}
\centerline{\epsfig{figure=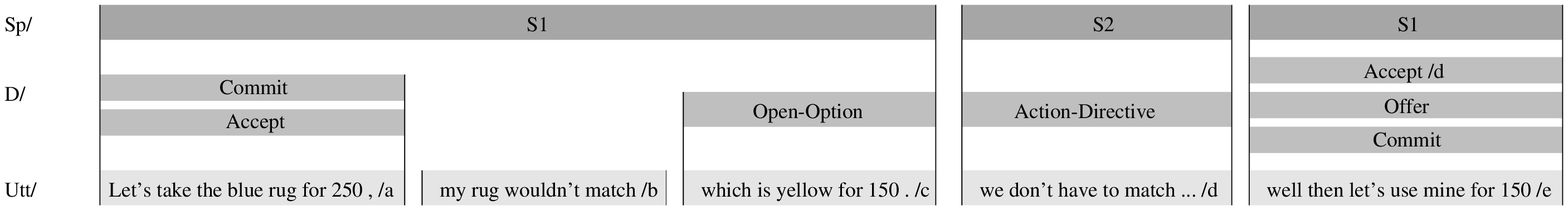,width=0.85\linewidth}}
\caption{Visualization of Annotation Graph for COCONUT Example}\label{fig:coconut}
\vspace*{2ex}\hrule
\end{figure*}

The COCONUT corpus is a set of dialogues in which
the two conversants collaborate on a task of deciding
what furniture to buy for a house \cite{DiEugenio98}.
The coding scheme
augments the DAMSL scheme \cite{AllenCore97} by having
some new top-level tags and by further specifying some existing
tags.  An example is given in Figure~\ref{fig:coconut-text}.

The example shows five utterance pieces, identified (a-e), four
produced by speaker S1 and one produced by speaker S2.
The discourse annotations can be glossed as follows:
\smtt{Accept} - the speaker is agreeing to a possible action or a claim;
\smtt{Commit} - the speaker potentially commits to intend to perform
a future specific action, and the commitment is not contingent
upon the assent of the addressee;
\smtt{Offer} - the speaker potentially commits to intend to perform
a future specific action, and the commitment is contingent
upon the assent of the addressee;
\smtt{Open-Option} - the speaker provides an option for the
addressee's future action;
\smtt{Action-Directive} - the utterance is designed to
cause the addressee to undertake a specific action.

In utterance (e) of Figure~\ref{fig:coconut-text},
speaker S1 simultaneously accepts
to the meta-action in (d) of not having matching colors,
and to the regular action of using S1's yellow rug.
The latter acceptance is not explicitly represented in the
original notation, so we shall only consider the former.

In representing this dialogue structure using annotation
graphs, we will be concerned to achieve the following:
(i) to treat multiple annotations of the same utterance fragment
as an unordered set, rather than a list, to simplify indexing
and query;
(ii) to explicitly link speaker S1 to utterances (a-c);
(iii) to formalize the relationship between \smtt{Accept(d)} and
utterance (d); and
(iv) formalize the rest of the annotation structure which is
implicit in the textual representation.

We adopt the types \smtt{Sp} (speaker),
\smtt{Utt} (utterance) and \smtt{D} (discourse).
A more refined type system could include other levels
of representation, it could distinguish forward versus backward
communicative function, and so on.
For the names we employ:
speaker identifiers \smtt{S1}, \smtt{S2};
discourse tags \smtt{Offer}, \smtt{Commit}, \smtt{Accept},
\smtt{Open-Option}, \smtt{Action-Directive}; and
orthographic strings representing the utterances.
For the classes (the third, optional field)
we employ the utterance identifiers
\smtt{a}, \smtt{b}, \smtt{c}, \smtt{d}, \smtt{e}.

An annotation graph representation of the COCONUT example can now
be represented as in Figure~\ref{fig:coconut}.
The arcs are structured into three layers, one for each type, where
the types are written on the left.  If the optional class field
is specified, this information follows the name field, separated by
a slash.  The \smtt{Accept/d} arc refers to the \smtt{S2} utterance
simply by virtue of the fact that both share the same class field.

Observe that the \smtt{Commit} and \smtt{Accept} tags for (a) are
unordered, unlike the original annotation.
and that speaker S1 is associated with all utterances
(a-c), rather than being explicitly linked to (a) and implicitly
linked to (b) and (c) as in Figure~\ref{fig:coconut-text}.

To make the referent of the \smtt{Accept} tag clear, we
make use of the class field.  Recall that the third component
of the fielded records, the class field, permits arcs to
refer to each other.  Both the referring and the referenced
arcs are assigned to equivalence class~\smtt{d}.

\subsection{Coreference Annotation in MUC-7}

\begin{figure*}
{\scriptsize\setlength{\tabcolsep}{.5\tabcolsep}
\begin{tabular}{l|l|l}
\begin{minipage}[t]{.3\linewidth}
\begin{alltt}
<COREF ID="2" MIN="woman">
  This woman
</COREF>
receives three hundred dollars a
month under 
<COREF ID="5">
  General Relief
</COREF>
, plus
<COREF ID="16"
       MIN="four hundred dollars">
  four hundred dollars a month in
  <COREF ID="17"
         MIN="benefits" REF="16">
    A.F.D.C. benefits
  </COREF>
</COREF>
for
<COREF ID="9" MIN="son">
  <COREF ID="3" REF="2">
    her
  </COREF>
  son
</COREF>
, who is
\end{alltt}
\end{minipage}
&
\begin{minipage}[t]{.3\linewidth}
\begin{alltt}
<COREF ID="10" MIN="citizen" REF="9">
  a U.S. citizen
</COREF>.
<COREF ID="4" REF="2">
  She
</COREF>
's among
<COREF ID="18" MIN="aliens">
  an estimated five hundred illegal
  aliens on
  <COREF ID="6" REF="5">
    General Relief
  </COREF>
  out of
  <COREF ID="11" MIN="population">
    <COREF ID="13" MIN="state">
      the state
    </COREF>
    's total illegal immigrant
    population of
    <COREF ID="12" REF="11">
      one hundred thousand
    </COREF>
  </COREF>
</COREF>
\end{alltt}
\end{minipage}
&
\begin{minipage}[t]{.3\linewidth}
\begin{alltt}
.
<COREF ID="7" REF="5">
  General Relief
</COREF>
is for needy families and unemployable
adults who don't qualify for other public
assistance.  Welfare Department spokeswoman
Michael Reganburg says
<COREF ID="15" MIN="state" REF="13">
  the state
</COREF>
will save about one million dollars a year if
<COREF ID="20" MIN="aliens" REF="18">
  illegal aliens
</COREF>
are denied
<COREF ID="8" REF="5">
  General Relief
</COREF>
.
\end{alltt}
\end{minipage}
\end{tabular}}

\caption{Coreference Annotation for BU Example}\label{fig:coref}
\vspace*{3ex}\hrule\vspace*{3ex}
\centerline{\epsfig{figure=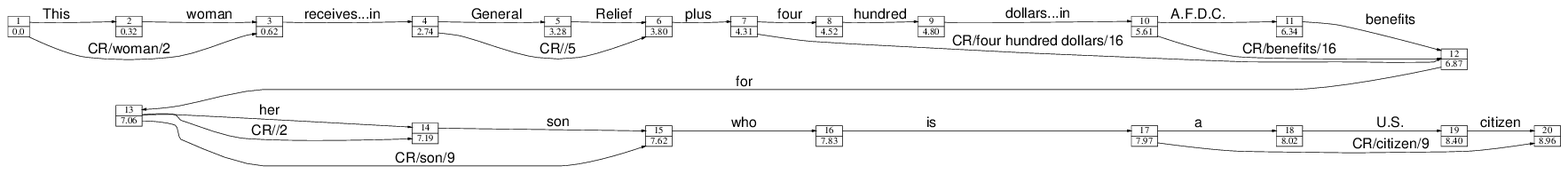,width=\linewidth}}
\caption{Annotation Graph for Coreference Example}\label{fig:coref-ag}
\vspace*{2ex}\hrule
\end{figure*}

The MUC-7 Message Understanding Conference specified
tasks for information extraction, named entity and coreference.
Coreferring expressions are to be linked
using SGML markup with \smtt{ID} and \smtt{REF} tags \cite{Hirschman97}.
Figure~\ref{fig:coref} is a sample of text from the
Boston University Radio Speech Corpus
[\smtt{www.ldc.upenn.edu/Catalog/LDC96S36.html}],
marked up with coreference tags.  (We are grateful to Lynette
Hirschman for providing us with this annotation.)

Noun phrases participating in coreference are wrapped with
\smtt{<coref>...</coref>} tags, which can bear the attributes
\smtt{ID}, \smtt{REF}, \smtt{TYPE} and \smtt{MIN}.  
Each such phrase is given a unique identifier, which may be
referenced by a \smtt{REF} attribute somewhere else.
Our example contains the following references:
$3 \rightarrow 2$,
$4 \rightarrow 2$,
$6 \rightarrow 5$,
$7 \rightarrow 5$,
$8 \rightarrow 5$,
$12 \rightarrow 11$,
$15 \rightarrow 13$.
The \smtt{TYPE} attribute encodes the relationship between the anaphor and
the antecedent.  Currently, only the identity relation is marked,
and so coreferences form an equivalence class.
Accordingly, our example contains the following equivalence classes:
$\left\{2, 3, 4\right\}$,
$\left\{5, 6, 7, 8\right\}$,
$\left\{11, 12\right\}$,
$\left\{13, 15\right\}$.

In our AG representation we choose the first number from each
of these sets as the identifier for the equivalence class.
MUC-7 also contains a specification for named entity annotation.
Figure~\ref{fig:bu-ne} gives an example, to be discussed
in \S\ref{sec:bu}.
This uses empty tags
to get around the problem of cross-cutting hierarchies.
This problem does not arise in the annotation graph formalism;
see \cite[2.7]{BirdLiberman99}.

\section{Hybrid Annotations}

There are many cases where a given
corpus is annotated at several levels,
from discourse to phonetics.
While a uniform structure is sometimes imposed,
as with Partitur \cite{Schiel98},
established practice and existing tools may
give rise to corpora transcribed using different
formats for different levels.  Two examples of hybrid
annotation will be discussed here: a TRAINS+DAMSL annotation,
and an eight-level annotation of the Boston University Radio
Speech Corpus.

\subsection{DAMSL annotation of TRAINS}

The TRAINS corpus \cite{Heeman93}
is a collection of about 100 dialogues containing
a total of 5,900 speaker turns
[\smtt{www.ldc.upenn.edu/Catalog} \smtt{/LDC95S25.html}].
Part of a transcript is shown below, where
\smtt{s} and \smtt{u} designate the two speakers,
\smtt{<sil>} denotes silent periods, and \smtt{+}
denotes boundaries of speaker overlaps.

\begin{bv}
utt1  : s:  hello <sil> can I help you 
utt2  : u:  yes <sil> um <sil> I have a problem here 
utt3  :     I need to transport one tanker of orange juice
            to Avon <sil> and a boxcar of bananas to
            Corning <sil> by three p.m. 
utt4  :     and I think it's midnight now 
utt5  : s:  uh right it's midnight 
utt6  : u:  okay so we need to <sil> 
            um get a tanker of OJ to Avon is the first
            thing we need to do 
utt7  :     + so + 
utt8  : s:  + okay + 
utt9  :     <click> so we have to make orange juice first 
utt10 : u:  mm-hm <sil> okay so we're gonna pick up <sil>
            an engine two <sil> from Elmira 
utt11 :     go to Corning <sil> pick up the tanker 
utt12 : s:  mm-hm 
utt13 : u:  go back to Elmira <sil> to get <sil> pick up
            the orange juice 
utt14 : s:  alright <sil> um well <sil> we also need to
            make the orange juice <sil> so we need to get
            + oranges <sil> to Elmira + 
utt15 : u:  + oh we need to pick up + oranges oh + okay + 
utt16 : s:  + yeah + 
utt17 : u:  alright so <sil> engine number two is going to
            pick up a boxcar 
\end{bv}

\begin{figure*}
{\scriptsize
\begin{minipage}[t]{.5\linewidth}
\begin{alltt}
This woman receives
<b_numex TYPE="MONEY">
  three hundred dollars
<e_numex>
a month under General Relief, plus
<b_numex TYPE="MONEY">
  four hundred dollars
<e_numex>
a month in A.F.D.C. benefits for her son, who is a
<b_enamex TYPE="LOCATION">
  U.S.
<e_enamex>
citizen.  She's among an estimated five hundred illegal
aliens on General Relief out of the state's total illegal
\end{alltt}
\end{minipage}
\begin{minipage}[t]{.5\linewidth}
\begin{alltt}
immigrant population of one hundred thousand.  General
Relief is for needy families and unemployable adults
who don't qualify for other public assistance.
<b_enamex TYPE="ORGANIZATION">
  Welfare Department
<e_enamex>
spokeswoman
<b_enamex TYPE="PERSON">
  Michael Reganburg
<e_enamex>
says the state will save about
<b_numex TYPE="MONEY">
  one million dollars
<e_numex>
a year if illegal aliens are denied General Relief. 
\end{alltt}
\end{minipage}}

\caption{Named Entity Annotation for BU Example}\label{fig:bu-ne}
\vspace*{2ex}\hrule
\end{figure*}

Accompanying this transcription are a number of xwaves label
files containing time-aligned word-level and segment-level
transcriptions.  Below, the start of file \smtt{speaker0.words}
is shown on the left, and the start of file \smtt{speaker0.phones}
is shown on the right.  The first number gives the file offset
(in seconds), and the middle number gives the label color.
The final part is a label for the interval which ends at the
specified time.  Silence is marked explicitly (again using \smtt{<sil>})
so we can infer that the first word `hello' occupies the
interval [0.110000, 0.488555].  Evidently the segment-level
annotation was done independently of the word-level annotation,
and so the times do not line up exactly.

\begin{bv}
 0.110000  122 <sil>             0.100000 122 <sil>
 0.488555  122 hello             0.220000 122 hh
 0.534001  122 <sil>             0.250000 122 eh ;*
 0.640000  122 can               0.330000 122 l
 0.690000  122 I                 0.460000 122 ow+1
 0.930000  122 help              0.530000 122 k
 1.068003  122 you               0.570000 122 ih
14.670000  122 <sil>             0.640000 122 n
14.920000  122 uh                0.690000 122 ay
15.188292  122 right             0.760000 122 hh
\end{bv}

The TRAINS annotations show the presence of
backchannel cues and overlap.  An example of
overlap is shown below:

\begin{bv}
50.130000  122 <sil>
50.260000  122 so
50.330000  122 we
50.480000  122 need
50.540000  122 to
50.651716  122 get
                                51.094197  122 <sil>
                                51.306658  122 oh
51.360000  122 oranges
                                51.410000  122 we
51.470000  122 <sil>
51.540000  122 to
                                51.560000  122 need
                                51.620000  122 to
                                51.850000  122 pick
51.975728  122 Elmira
                                52.020000  122 up
                                52.470000  122 oranges
                                52.666781  122 oh
52.807837   76 <sil>
                                52.940000  122 okay
53.047996   76 yeah
                                53.535600  122 <sil>
                                53.785600  122 alright
                                54.303529  122 so
\end{bv}

\begin{figure*}
\begin{sv}
<Dialog Id=d92a-2.2 Annotation-date="08-14-97" Annotator="Reconciled Version"
  Speech="/d92a-2.2/dialog.fea" Status=Verified>
...
<Turn Id=T9 Speaker="s" Speech="-s 44.853889 -e 52.175728">
...
<Utt Id=utt17 Agreement=None Influence-on-listener=Action-directive Influence-on-speaker=Commit Info-level=Task Response-to=""
  Speech="-s 45.87 -e 52.175728" Statement=Assert>
[sil] um well [sil] we also need to make the orange juice [sil]
so we need to get + oranges [sil] to Elmira +
<Turn Id=T10 Speaker="u" Speech="-s 51.106658 -e 53.14">
<Utt Id=utt18 Agreement=Accept Influence-on-listener=Action-directive Influence-on-speaker=Commit Info-level=Task
  Response-to="utt17" Speech="-s 51.106658 -e 52.67" Statement=Assert Understanding=SU-Acknowledge>
+ oh we need to pick up + oranges
<Utt Id=utt19 Agreement=Accept Influence-on-speaker=Commit Info-level=Task Response-to="utt17" Speech="-s 52.466781 -e 53.14"
  Understanding=None>
oh + okay +
<Turn Id=T11 Speaker="s" Speech="-s 52.047996 -e 53.247996">
<Utt Id=utt20 Agreement=Accept Info-level=Task Response-to="utt18" Speech="-s 52.047996 -e 53.247996" Understanding=SU-Acknowledge>
+ yeah +
...
</Dialog>
\end{sv}
\caption{DAMSL Annotation of a TRAINS Dialogue}\label{fig:damsl} 
\vspace*{2ex}\hrule
\end{figure*}

As seen in Figure~\ref{fig:ldc2} and explained more fully
in \cite{BirdLiberman99}, overlap carries no implications
for the internal structure of speaker turns or for the
position of turn-boundaries.

Now, independently of this annotation there is also a dialogue
annotation in DAMSL, as shown in Figure~\ref{fig:damsl}.
Here, a dialog is broken down into turns and thence into utterances,
where the tags contain discourse-level annotation.

\begin{figure*}
\centerline{\epsfig{figure=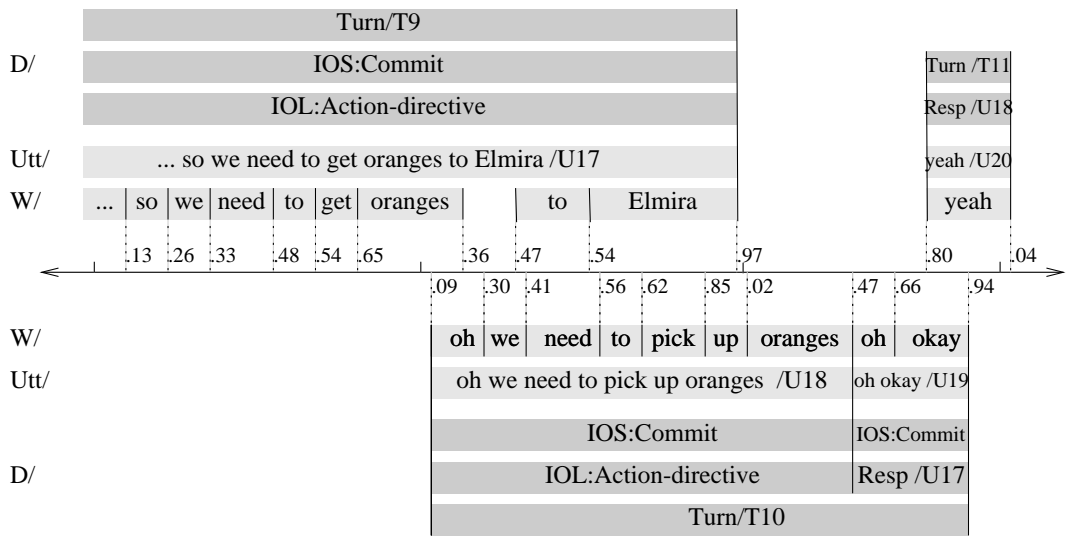,width=.85\linewidth}}
\caption{Graph Structure for TRAINS Example}\label{fig:trains}
\vspace*{2ex}\hrule
\end{figure*}

In representing this hybrid annotation as an AG
we are motivated by the following concerns.
First, we want to preserve the distinction between the TRAINS
and DAMSL components, so that they can remain in their native
formats (and be manipulated by their native tools) and be
converted independently to AGs then combined using AG union,
and so that they can be projected back out if necessary.
Second, we want to identify those boundaries that necessarily have the
same time reference (such as the end of utterance 17 and the
end of the word `Elmira'), and represent them using a single
graph node.  Contributions from different speakers will remain
disconnected in the graph structure.
Finally, we want to use the equivalence class names to allow
cross-references between utterances.
A fragment of the proposed annotation graph is depicted using
our visualization format in Figure~\ref{fig:trains}.
Observe that, for brevity, some discourse tags are not represented,
and the phonetic segment level is omitted.

Note that the tags in Figure~\ref{fig:damsl} have the form
of fielded records and so, according to the AG definition,
all the attributes of a tag could be put into a single label.
We have chosen to maximally split such records into multiple
arc labels, so that search predicates do not need to take
account of internal structure, and to limit the consequences
of an erroneous code.  A relevant analogy here is that
of pre-composed versus compound characters in
Unicode.  The presence of both forms of a character in a text
raises problems for searching and collating.  This problem
is avoided through normalization, and this is typically done
by maximally decomposing the characters.

\subsection{Multiple annotations of the BU corpus}
\label{sec:bu}

Linguistic analysis is always multivocal, in two senses. First, there
are many types of entities and relations, on many scales, from
acoustic features spanning a hundredth of a second to narrative
structures spanning tens of minutes. Second, there are many
alternative representations or construals of a given kind of
linguistic information.

Sometimes these alternatives are simply more or less convenient for a
certain purpose.  Thus a researcher who thinks theoretically of
phonological features organized into moras, syllables and feet, will
often find it convenient to use a phonemic string as a
representational approximation. In other cases, however, different
sorts of transcription or annotation reflect different theories about
the ontology of linguistic structure or the functional categories of
communication.

The AG representation offers a way to deal productively with both
kinds of multivocality. It provides a framework for relating different
categories of linguistic analysis, and at the same time to compare
different approaches to a given type of analysis. 

\begin{figure*}
\centerline{\epsfig{figure=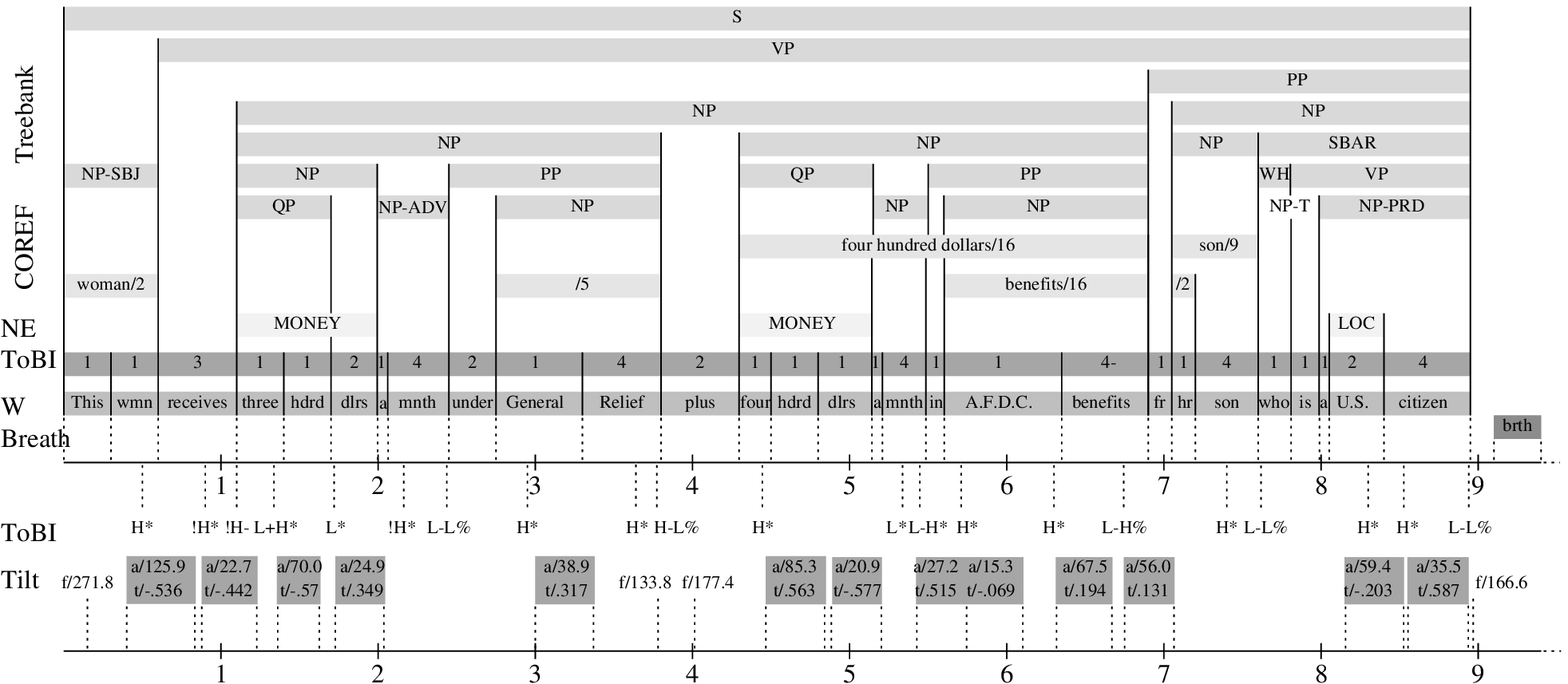,width=\linewidth}}
\caption{Visualization for BU Example}\label{fig:bu}
\vspace*{2ex}\hrule
\end{figure*}

As an example, Figure~\ref{fig:bu} shows an AG-based visualization of eight
different sorts of annotation of a phrase from the BU Radio Corpus,
produced by Mari Ostendorf and others at Boston University, and
published by the LDC
[\smtt{www.ldc.upenn.edu/Catalog/LDC96S36.html}].
The basic material is from a recording of
a local public radio news broadcast. The BU annotations include four
types of information: orthographic transcripts, broad phonetic
transcripts (including main word stress), and two kinds of prosodic
annotation, all time-aligned to the digital audio files. The two kinds
of prosodic annotation implement the system known as ToBI
[\smtt{www.ling.ohio-state.edu/phonetics/E\_ToBI/}].
ToBI is an acronym for ``Tones and Break Indices'', and correspondingly
provides two types of information: {\em Tones}, which are taken from a
fixed vocabulary of categories of (stress-linked) ``pitch accents'' and
(juncture-linked) ``boundary tones''; and {\em Break Indices}, which are
integers characterizing the strength and nature of interword
disjunctures.

We have added four additional annotations: coreference annotation and
named entity annotation in the style of MUC-7
[\smtt{www.muc.saic.com/proceedings/muc\_7\_toc.html}] provided
by Lynette Hirschman;
syntactic structures in the style of the Penn TreeBank \cite{Marcus93}
provided by Ann Taylor;
and an alternative
annotation for the F$_0$ aspects of prosody, known as {\em Tilt}
\cite{Taylor98tilt} and
provided by its inventor, Paul Taylor. Taylor has done Tilt annotations
for much of the BU corpus, and will soon be publishing them as a point of
comparison with the ToBI tonal annotation. Tilt differs from ToBI in
providing a quantitative rather than qualitative characterization
of F$_0$ obtrusions: where ToBI might say ``this is a L+H* pitch accent,''
Tilt would say ``This is an F$_0$ obtrusion that starts at time $t_0$,
lasts for duration $d$ seconds, involves $a$ Hz total F$_0$ change,
and ends $l$ Hz different in F$_0$ from where it started.''

As usual, the various annotations come in a bewildering variety of
file formats. These are not entirely trivial to put into registration,
because (for instance) the TreeBank terminal string contains both more
(e.g.\ traces) and fewer (e.g.\ breaths) tokens than the orthographic
transcription does. One other slightly tricky point: the connection
between the word string and the ``break indices'' (which are ToBI's
characterizations of the nature of interword disjuncture) are mediated
only by identity in the floating-point time values assigned to word
boundaries and to break indices in separate files. Since these time
values are expressed as ASCII strings, it is easy to lose the identity
relationship without meaning to, simply by reading in and writing out
the values to programs that may make different choices of internal
variable type (e.g.\ float vs.\ double), or number of decimal digits to
print out, etc.

Problems of this type are normal whenever multiple annotations need to
be compared. Solving them is not rocket science, but does take careful
work.  When annotations with separate histories involve mutually
inconsistent corrections, silent omissions of problematic material, or
other typical developments, the problems are multiplied.
In noting such difficulties, we are not criticizing the authors of the
annotations, but rather observing the value of being able to put
multiple annotations into a common framework.

Once this common framework is established, via translation of all eight
``strands'' into AG graph terms, we have the basis for posing queries
that cut across the different types of annotation.
For instance, we might look at the distribution of Tilt parameters
as a function of ToBI accent type; or the distribution of Tilt
and ToBI values for initial vs. non-initial members of coreference
sets; or the relative size of Tilt F0-change measures for nouns vs.
verbs.

We do not have the space in this paper to discuss the design of an
AG-based query formalism at length -- and indeed, many details of
practical AG query systems remain to be decided -- but a short
discussion will indicate the direction we propose to take. Of course the
crux is simply to be able to put all the different annotations into
the same frame of reference, but beyond this, there are some aspects
of the annotation graph formalism that have nice properties for
defining a query system. For example, if an annotation graph is
defined as a set of ``arcs'' like those given in the XML encoding in
\S\ref{sec:intro},
then every member of the power set of this arc set is also
a well-formed annotation graph. The power set construction provides
the basis for a useful query algebra, since it defines the complete
set of possible values for queries over the AG in question, and is
obviously closed under intersection, union and relative complement.
As another example, various time-based indexes are definable on an
adequately time-anchored annotation graph, with the result that many
sorts of precedence, inclusion and overlap relations are easy to
calculate for arbitrary subgraphs.  See \cite[\S 5]{BirdLiberman99}
for discussion.

In this section, we have indicated some of the ways in which the
AG framework can facilitate the analysis of complex combinations
linguistic annotations. These annotation sets are typically
multivocal, both in the sense of covering multiple types of linguistic
information, and also in the sense of providing multiple versions of
particular types of analysis. Discourse studies are especially multivocal
in both senses, and so we feel that this approach will be especially
helpful to discourse researchers.

\section{Conclusion}

This proliferation of formats and approaches can be viewed as a sign
of intellectual ferment.  The fact that so many people have devoted so
much energy to fielding new entries into this bazaar of data formats
indicates how important the computational study of communicative
interaction has become.  However, for many researchers, this
multiplicity of approaches has produced headaches and confusion,
rather than productive scientific advances.  We need a way to
integrate these approaches without imposing some form of premature
closure that would crush experimentation and innovation.

Both here, and in associated work \cite{BirdLiberman99},
we have endeavored to show how all current annotation formats
involve the basic actions of associating labels with stretches of
recorded signal data, and attributing logical sequence, hierarchy
and coindexing to such labels.
We have grounded this assertion by defining
annotation graphs and by showing how a disparate range of annotation
formats can be mapped into AGs.  This work provides a central piece
of the algebraic foundation for inter-translatable formats and
inter-operating tools.  The intention is not to replace the formats
and tools that have been accepted by any existing community of
practice, but rather to make the descriptive and analytical practices,
the formats, data and tools universally accessible.
This means that
annotation content for diverse domains and theoretical
models can be created
and maintained using tools that are the most suitable or familiar
to the community in question.  It also means that
we can get started on integrating annotations, corpora and
research findings right away, without having to wait
until final agreement on all possible tags and attributes has been
achieved.

There are many existing approaches to discourse annotation, and
many options for future approaches.  Our explorations
presuppose a particular set of goals:
(i) generality, specificity, simplicity;
(ii) searchability and browsability; and
(iii) maintainability and durability.
These are discussed in full in \cite[\S 6]{BirdLiberman99}.
By identifying a common conceptual core
to all annotation structures,
we hope to provide a foundation for a
wide-ranging integration of tools, formats and corpora.
One might, by analogy to translation systems, describe AGs
as an interlingua which permits free exchange of annotation data
between $n$ systems once $n$ interfaces have been written, rather
than $n^2$ interfaces.

Although we have been primarily concerned with the
structure rather than the content of annotations,
the approach opens the way to meaningful evaluation
of content and comparison of contentful differences
between annotations, since it is possible to do
all manner of quasi-correlational analyses of parallel annotations.
A tool for converting a given format into the
AG framework only needs to be written once.
Once this has been done, it becomes a straightforward task to
pose complex queries over multiple corpora.
Whereas if one
were to start with annotations in several distinct file formats,
it would be a major programming chore to ask even a simple question.

\section*{Acknowledgements}

We are grateful to the following people for discussions and
input concerning the material presented here:
Chris Cieri,
Dave Graff,
Julia Hirschberg,
Lynette Hirschman,
Brian MacWhinney,
Ann Taylor,
Paul Taylor,
Marilyn Walker,
and three anonymous reviewers.

\raggedright

\end{document}